\documentclass[conference]{IEEEtran}
\IEEEoverridecommandlockouts
\usepackage{cite}
\usepackage{amsmath,amssymb,amsfonts}
\usepackage{algorithmic}
\usepackage{graphicx}
\usepackage{textcomp}
\usepackage[colorlinks,linkcolor=red]{hyperref}

\usepackage{multirow}
\usepackage[table,xcdraw]{xcolor}
\usepackage{booktabs}

\makeatletter
\newcommand{\linebreakand}{%
  \end{@IEEEauthorhalign}
  \hfill\mbox{}\par
  \mbox{}\hfill\begin{@IEEEauthorhalign}
}
\makeatother

\def\BibTeX{{\rm B\kern-.05em{\sc i\kern-.025em b}\kern-.08em
    T\kern-.1667em\lower.7ex\hbox{E}\kern-.125emX}}
\begin{document}

\title{ABC: Attention with Bilinear Correlation for Infrared Small Target Detection}



\author{

\IEEEauthorblockN{Peiwen Pan}
\IEEEauthorblockA{\textit{School of Computer Science and Engineering} \\
\textit{Nanjing University of Science and Technology}\\
Nanjing, China \\
121106022690@njust.edu.cn}
\and
\IEEEauthorblockN{Huan Wang$^*$\thanks{*Corresponding author}}
\IEEEauthorblockA{\textit{School of Computer Science and Engineering} \\
\textit{Nanjing University of Science and Technology}\\
Nanjing, China \\
wanghuanphd@njust.edu.cn}
\linebreakand
\IEEEauthorblockN{Chenyi Wang}
\IEEEauthorblockA{\textit{School of Computer Science and Engineering} \\
\textit{Nanjing University of Science and Technology}\\
Nanjing, China \\
wcyjerry@njust.edu.cn}
\and
\IEEEauthorblockN{Chang Nie}
\IEEEauthorblockA{\textit{School of Computer Science and Engineering} \\
\textit{Nanjing University of Science and Technology}\\
Nanjing, China \\
changnie@njust.edu.cn}
}

\maketitle

\begin{abstract}
Infrared small target detection (ISTD) has a wide range of applications in early warning, rescue, and guidance. However, CNN based deep learning methods are not effective at segmenting infrared small target (IRST) that it lack of clear contour and texture features, and transformer based methods also struggle to achieve significant results due to the absence of convolution induction bias. To address these issues, we propose a new model called attention with bilinear correlation (ABC), which is based on the transformer architecture and includes a convolution linear fusion transformer (CLFT) module with a novel attention mechanism for feature extraction and fusion, which effectively enhances target features and suppresses noise. Additionally, our model includes a u-shaped convolution-dilated convolution (UCDC) module located deeper layers of the network, which takes advantage of the smaller resolution of deeper features to obtain finer semantic information. Experimental results on public datasets demonstrate that our approach achieves state-of-the-art performance. Code is available at \url{https://github.com/PANPEIWEN/ABC}
\end{abstract}


\section{Introduction}
\label{sec:intro}
\begin{figure}
    \centering
    \includegraphics[width=1\linewidth]{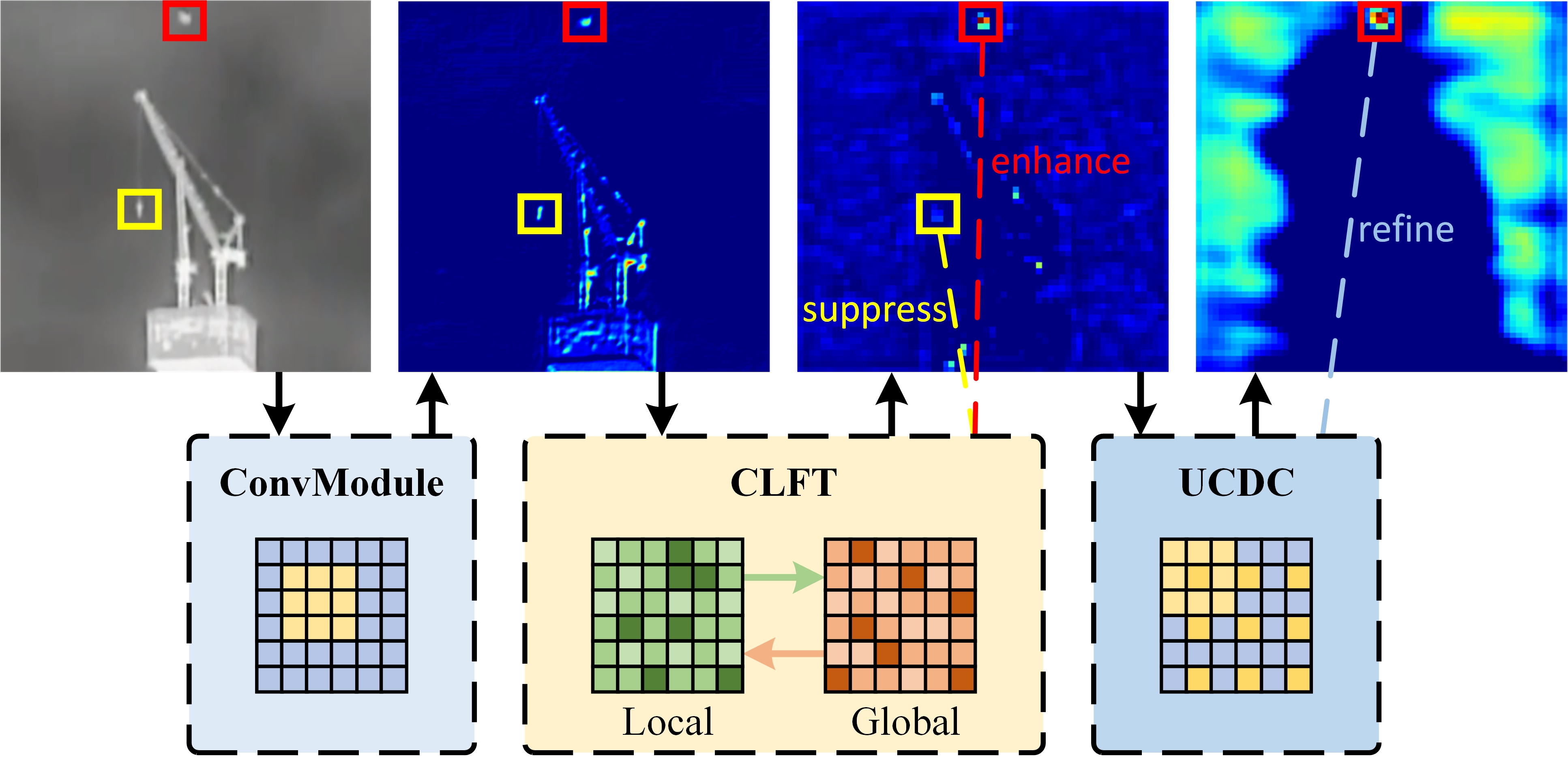}
    \caption{Feature map visualization of infrared image at different stages. The red box is IRST, and the yellow box is noise. The first picture is the original picture, and the second picture is the feature map after a layer of the convolution module, there is obvious noise interference. The third picture is the feature map after the CLFT modules, which suppresses the noise well and enhances the target features. The fourth picture is the result after passing through the UCDC module, which processes the target features in a more refined manner.}
    \label{fig1}
\end{figure}

Infrared small target detection (ISTD) has numerous applications in maritime surveillance, early warning systems, precision guidance, and disaster relief. As shown in Fig. \ref{fig1}, compared with common visible target, infrared small target (IRST) has the following characteristics: 1) Due to the long imaging distance, the pixel ratio of IRST pixels to the whole image is very small. 2) The energy of infrared radiation is significantly attenuated over distance, making it easy for IRST to drown in background clutter and sensor noise. 3) IRST is very sparse, leading to a severe imbalance between object and background regions. This makes detecting IRST very challenging.

A large number of model-driven based traditional methods have been proposed to be able to effectively detect IRST. The filter-based method \cite{deshpande1999max,zhao2004background,marvasti2018flying,anju2016shearlet,wang2017infrared} mainly uses the designed filter to estimate the infrared image background to achieve the effect of suppressing the background. The local contrast-based method \cite{mazzu2016cognitive,chen2016efficient,deng2016small,deng2017entropy,han2019local} uses local differences to suppress the background to enhance the target. The low-rank and sparse matrix recovery-based method \cite{gao2013infrared,kong2021infrared,zhu2019infrared,dai2017reweighted,zhang2018infrared,zhong2023infrared} uses the sparse property of IRST and the low-rank property of the background to turn the detection task into a classification task. However, these methods often rely heavily on a priori knowledge and are sensitive to hyperparameters, and not perform well on images with complex backgrounds and noise.

With the advancement of deep learning, numerous data-driven based deep learning methods have been proposed. MDvsFA \cite{wang2019miss} uses generative adversarial networks to balance missed detections and false alarms. ACM \cite{dai2021asymmetric} proposes an asymmetric context modulation fusion module to fuse deep features with shallow features. ISNet \cite{zhang2022isnet} uses Taylor finite difference and bi-directional attention aggregation blocks to accurately detect the shape features of IRST. However, these methods are limited by the local receptive field of the convolution operation, lack the ability of global information perception, they may also be prone to detecting noise in the infrared image as the target. And because the IRST is small, it is easy to cause target loss after multiple downsampling operations, and this process is irreversible.

Transformer \cite{vaswani2017attention, dosovitskiy2020image} structure has excellent global feature characterization capability, but it may not be effective at detecting IRST, which has few distinctive features. Therefore, we effectively combine the local correlation of the CNN and the global correlation of the transformer to propose a new model: ABC. The overall ABC is an encoder-decoder structure similar to UNet \cite{ronneberger2015u}, in which the encoder is composed of convolution module and convolution linear fusion transformer (CLFT) module, and the decoder is composed of u-shaped convolution-dilated convolution (UCDC) module and convolution module. For the CLFT module, we use the transformer structure, redesign the self attention architecture, and introduce convolution and dilated convolution. Calculate the attention matrix through the bilinear attention module (BAM), and fuse it with the features extracted by the convolution operation. The output is then obtained through a feedforward layer. The CLFT module obtains both local and global features, which can effectively enhances the target features and suppresses the noise, effectively avoiding the problem of losing IRST when the network is too deep. The UCDC module is a u-shaped structure consisting of convolution and dilated convolution layers, and located in the deeper layers of the network. It is able to extract finer features from the feature map which has been downsampled several times and has a smaller resolution.

In summary, the main contributions of this paper are as follows:

1) The CLFT module designed based on transformer structure can effectively enhance target features and suppress noise.

2) The UCDC module makes full use of the characteristics of deep features and can process the deep features of the network more finely.

3) The proposed method achieves state-of-the-art performance on all existing public datasets.

\section{The proposed method}

\begin{figure*}[h]
    \centering
    \includegraphics[width=1.0\linewidth]{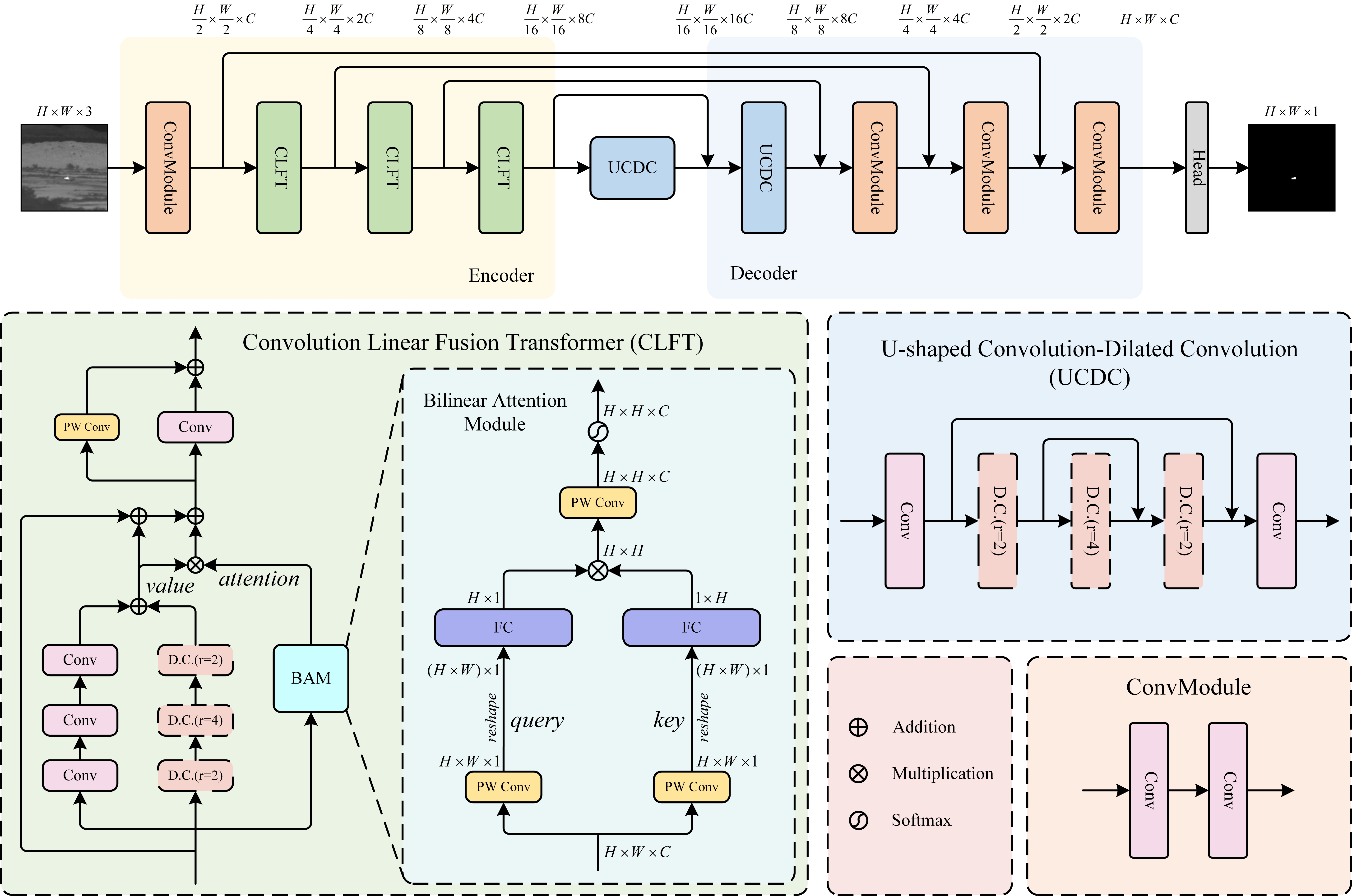}
    \caption{Overview of the proposed ABC. ABC is an encoder-decoder structure similar to UNet, which consists of convolution modules, convolution linear fusion transformer (CLFT) modules, and u-shaped convolution-dilated convolution (UCDC) modules . The convolution module consists of two convolution layers. CLFT module is a transformer structure consisting of convolution layers, dilated convolution layers, pointwise convolution layers, and a bilinear attention module (BAM). Where BAM consists of pointwise convolution layers and fully connected layers, through which the attention matrix is calculated. UCDC module is a u-shaped structure consisting of convolution layers and dilated convolution layers.}
    \label{fig2}
\end{figure*}

\subsection{Overall Architecture}
The general architecture of the ABC model is shown in Fig. \ref{fig2}. It is an encoder-decoder structure similar to UNet, with the encoder consisting of one convolution module and three CLFT modules and the decoder consisting of one UCDC module and three convolution modules. The convolution module consists of two ordinary convolution layers. There is also a UCDC module serving as a transition layer between the encoder and decoder and a pointwise convolution layer behind the decoder as a segmentation head to output the segmentation result. The model uses skip connections between the encoder and decoder to achieve cross-layer feature fusion.

\subsection{Convolution Linear Fusion Transformer}
\label{sec3_2}
The traditional CNN based model lacks global modeling ability and can only extract features locally, then it is easily disturbed by noise. And after multiple downsampling, IRST is easily lost in the deeper layers of the network. The advantage of the transformer is its global modeling capability, but due to the specificity of IRST, local feature extraction is also essential if we want to achieve good results in this task. For this reason, we rethought the relationship between local and global, CNN and transformer, and designed the CLFT module based on the transformer structure. As shown in Fig. \ref{fig2}, this module redesigns the self-attention architecture, which mainly consists of convolution layers, dilated convolution layers, and fully connected layers. We will introduce the CLFT module in detail next.

First, we introduce the bilinear attention module (BAM), which is responsible for computing the attention matrix. As shown in Fig. \ref{fig2}, the BAM consists of pointwise convolution layers and fully connected layers. Given an input feature ${\bf I}\in \mathbb{R}^{H\times W\times C}$, ${\bf I}_1\in \mathbb{R}^{H\times W\times 1}$ and ${\bf I}_2\in \mathbb{R}^{H\times W\times 1}$ are obtained after two pointwise convolution layers, respectively. Then reshape ${\bf I}_1$ and ${\bf I}_2$ to obtain $q\in \mathbb{R}^{(H\times W)\times 1}$ and $k\in \mathbb{R}^{(H\times W)\times 1}$, and then $q$ and $k$ are calculated and reshaped through the fully connected layer to obtain $q\in \mathbb{R}^{H\times 1}$ and $k\in \mathbb{R}^{1\times H}$. Then perform matrix multiplication on $q$ and $k$ to get the $attention\in \mathbb{R}^{H\times H}$. Finally, perform pointwise convolution and softmax operations on the $attention$ to obtain the output $attention\in \mathbb{R}^{H\times H\times C}$. The specific calculation process is shown in \eqref{eq1}.
\begin{equation}
\label{eq1}
\setlength{\abovedisplayskip}{3pt} 
\setlength{\belowdisplayskip}{3pt}
\renewcommand{\arraystretch}{1.3}
    \begin{array}{cc} 
        \displaystyle \displaystyle q={\rm FC}(reshape({\rm PW}({\bf I})))\\
        \displaystyle \displaystyle k={\rm FC}(reshape({\rm PW}({\bf I})))\\
        \displaystyle \displaystyle attention=softmax({\rm PW}(q\times k))
    \end{array}
\end{equation}
Where ${\rm PW}(\cdot)$ denotes pointwise convolution and ${\rm FC}(\cdot)$ denotes fully connected layer. The attention matrix can effectively perceive the position of the target in the feature map after continuous optimization learning.

The $v$ in CLFT module is calculated by three convolution layers and three dilated convolution layers with dilation rates of 2, 4, and 2, respectively. Where the convolution layer has a small receptive field and can effectively extract target features. The receptive field of the dilated convolution layer is larger than that of the convolution layer, which can obtain long-distance information and be used to compensate for the shortcomings of the convolution layer in that the features extracted are not fine enough due to the small receptive field. Although the sparse sampling method of the dilated convolution will cause some pixels not to be utilized, which will destroy the continuity and correlation of the information, this drawback can be well avoided due to the non-local autocorrelation of the background of the infrared image.  Convolution layers and dilated convolution layers complement each other and make up for shortcomings mutually. The combination of the two can effectively obtain both near and far information and extract finer features.

Given the input feature ${\bf I}\in \mathbb{R}^{H\times W\times C}$, we can obtain ${\bf I}_{conv}\in \mathbb{R}^{H\times W\times C}$ and ${\bf I}_{dconv}\in \mathbb{R}^{H\times W\times C}$ by calculating through the convolution layers and the dilated convolution layers, respectively, and then ${\bf I}_{conv}$ and ${\bf I}_{dconv}$ are added to obtain $v\in \mathbb{R}^{H\times W\times C}$. Finally, perform matrix multiplication of $attention$ and $v$ to get the output ${\bf O}_{attention}\in \mathbb{R}^{H\times W\times C}$. The specific calculation process is shown in \eqref{eq2}.
\begin{equation}
\label{eq2}
\setlength{\abovedisplayskip}{3pt} 
\setlength{\belowdisplayskip}{3pt}
\renewcommand{\arraystretch}{1.3}
    \begin{array}{cc} 
        \displaystyle \displaystyle v={\rm Conv}({\bf I})+{\rm DConv}({\bf I})\\
        \displaystyle \displaystyle {\bf O}_{attention}=attention\times v
    \end{array}
\end{equation}
Where ${\rm Conv}(\cdot)$ denotes convolution and ${\rm DConv}(\cdot)$ denotes dilated convolution. The $v$ is the feature matrix extracted by convolution and dilate convolution layers, and may be prone to extracting noise as IRST due to the lack of global information. However, after global modeling using the $attention$, the model can focus on IRST. The resulting feature matrix, which is obtained through matrix multiplication, is endowed with global information, which helps guide the extraction of target features.

Finally, we fuse the extracted local features and global features and then calculate the final output ${\bf O}\in \mathbb{R}^{H\times W\times 2C}$ of this module through a feedforward layer consisting of a convolution layer and a pointwise convolution layer. The specific calculation process is shown in \eqref{eq3}.
\begin{equation}
\label{eq3}
\setlength{\abovedisplayskip}{3pt} 
\setlength{\belowdisplayskip}{3pt}
\renewcommand{\arraystretch}{1.5}
    \begin{array}{cc} 
        \displaystyle \displaystyle \hat{{\bf O}}={\bf I}+v+\alpha {\bf O}_{attention}\\
        \displaystyle \displaystyle {\bf O}={\rm Conv}(\hat{{\bf O}})+{\rm PW}(\hat{{\bf O}})
    \end{array}
\end{equation}
Where $\alpha$ is the learnable weight, ${\rm Conv}(\cdot)$ denotes convolution, ${\rm PW}(\cdot)$ denotes pointwise convolution, and $\hat{{\bf O}}\in \mathbb{R}^{H\times W\times C}$ is the output after feature fusion. Feature fusion can supplement the semantic information lost due to assigning local features to global information.  By stacking three CLFT modules layer by layer, the model is able to effectively enhance target features and suppress noise, and it is able to address the problem of target loss during downsampling.
\subsection{U-shaped Convolution-Dilated Convolution}
\label{sec3_3}
After the input feature map is processed by the CLFT modules, most of the noise has been filtered out and the target features are more significant when it is passed to the deeper layers of the network. At this point, the feature map has a small resolution and the receptive field of the convolution operation is relatively large. To handle these deep features more finely, we design the UCDC module.

As shown in Fig.~\ref{fig2}, the UCDC module is a u-shaped structure consisting of two convolution layers and three dilated convolution layers with dilation rates of 2, 4, and 2. Since the deep features contain less semantic information, we use skip connections for feature compensation to prevent information loss due to convolution operations. 

The UCDC module makes full use of the characteristics of the deep features of the network.  The first three layers have progressively larger receptive fields, which allows for the processing of a larger range of pixels and effectively filters out the noise around IRST. Due to the small resolution of the feature map, the target is extremely small in the feature map at this time after the first three layers of processing. Therefore, we gradually make the receptive field smaller and use a small receptive field to make the extracted features more detailed. After processing by the UCDC module, we can obtain a clean and fine feature map.

\section{Experiment}

\begin{table*}[ht]
\centering
\caption{$IoU(\%)$, $nIoU(\%)$, $F_1(10^{-2})$ of different SOTA methods on NUAA, IRSTD1k, SIRSTAUG, and NUDT datasets.}
\tabcolsep=0.23cm
\renewcommand\arraystretch{1.5}
\begin{center}
\begin{tabular}{cccccccccccccccc}
\toprule[1pt]
                        & \multicolumn{3}{c}{NUAA}                                                                                               &                                  & \multicolumn{3}{c}{IRSTD1k}                                                                                            &                                  & \multicolumn{3}{c}{SIRSTAUG}                                                                                           &                                  & \multicolumn{3}{c}{NUDT}                                                                                               \\ \cline{2-4} \cline{6-8} \cline{10-12} \cline{14-16} 
\multirow{-2}{*}{Model} & IoU $\uparrow$                                   & nIoU $\uparrow$                                 & $\rm F_1$ $\uparrow$                                     &                                  & IoU $\uparrow$                                   & nIoU $\uparrow$                                  & $\rm F_1$ $\uparrow$                                     &                                  & IoU $\uparrow$                                   & nIoU $\uparrow$                                  & $\rm F_1$ $\uparrow$                                     &                                  & IoU $\uparrow$                                   & nIoU $\uparrow$                                  & $\rm F_1$ $\uparrow$                                     \\ \hline
IPI \cite{gao2013infrared}                     &57.64                                       &63.74                                       &73.12                                        &                                  &14.98                                       &34.51                                       &26.05                                        &                                  &37.75                                        &45.29                                        &54.80                                         &                                  &37.49                                       &48.38                                       &54.53                                        \\
RIPT \cite{dai2017reweighted}                    &28.38                                       &35.91                                       &44.21                                        &                                  &11.33                                       &17.43                                       &20.35                                        &                                  &24.13                                       &33.98                                       &38.88                                        &                                  &29.17                                      &36.12                                       &45.16                                        \\
PSTNN \cite{zhang2019infrared}                   &51.52                                       &61.92                                       &68.00                                        &                                  &15.93                                       &32.71                                       &27.48                                        &                                  &19.14                                       &27.16                                       &32.13                                        &                                  &27.72                                       &39.80                                       &43.41                                        \\
ACM \cite{dai2021asymmetric}                     & 72.88                                 & 72.17                                 & 84.31                                 &                                  & 63.39                                 & 60.81                                 & 77.59                                 &                                  & 73.84                                 & 69.83                                 & 84.95                                 &                                  & 68.48                                 & 69.26                                 & 81.29                                 \\
AGPCNet \cite{zhang2021agpcnet}                 & 77.13                                 & 75.19                                 & 87.09                                 &                                  & 68.81                                 & 66.18                                 & 81.52                                 &                                  & 74.71                                 & 71.49                                 & 85.52                                 &                                  & 88.71                                 & 87.48                                 & 94.02                                 \\
DNANet \cite{li2022dense}                  & 74.91                                 & 75.11                                 & 85.66                                 &                                  & 68.87                                 & 67.53                                 & 81.57                                 &                                  & 74.88                                 & 70.23                                 & 85.64                                 &                                  & 92.67                                 & 92.09                                 & 96.20                                  \\

RKFormer \cite{zhang2022rkformer}                & 77.24                                 & 74.89                                 & 87.15                                      &                                  & 64.12                                 & 64.18                                 & 78.13                                      &                                  & -                                     & -                                     & -                                      &                                  & -                                     & -                                     & -                                      \\
ISNet \cite{zhang2022isnet}                   & 80.02                                 & 78.12                                 & 88.90                                      &                                  & 68.77                                 & 64.84                                 & 81.49                                      &                                  & -                                     & -                                     & -                                      &                                  & -                                     & -                                     & -                                      \\ 
ABC(ours)              & {\color[HTML]{FE0000} \textbf{81.01}} & {\color[HTML]{FE0000} \textbf{79.00}} & {\color[HTML]{FE0000} \textbf{89.51}} & {\color[HTML]{FE0000} \textbf{}} & {\color[HTML]{FE0000} \textbf{72.02}} & {\color[HTML]{FE0000} \textbf{68.81}} & {\color[HTML]{FE0000} \textbf{83.73}} & {\color[HTML]{FE0000} \textbf{}} & {\color[HTML]{FE0000} \textbf{76.12}} & {\color[HTML]{FE0000} \textbf{71.83}} & {\color[HTML]{FE0000} \textbf{86.44}} & {\color[HTML]{FE0000} \textbf{}} & {\color[HTML]{FE0000} \textbf{92.85}} & {\color[HTML]{FE0000} \textbf{92.45}} & {\color[HTML]{FE0000} \textbf{96.29}} \\ \bottomrule[1pt]
\end{tabular}
\end{center}
\label{tab1}
\end{table*}

\subsection{Datasets and Evaluation Metrics}
\textbf{Datasets:} We conducted experiments on the public available NUAA \cite{dai2021asymmetric}, IRSTD1k \cite{zhang2022isnet}, SIRSTAUG \cite{zhang2021agpcnet}, and NUDT \cite{li2022dense} datasets. For each dataset, we used 80\% of the images as the training set and 20\% of the images as the test set.

\textbf{Evaluation Metrics:} We used intersection over union ($IoU$), normalized intersection over union ($nIoU$), and F1 score ($F_1$) as evaluation metrics for our experiment. They are defined as:
\begin{equation}
\label{eq4}
\setlength{\abovedisplayskip}{3pt} 
\setlength{\belowdisplayskip}{3pt}
\renewcommand{\arraystretch}{2.5}
    \begin{array}{cc} 
        \displaystyle \displaystyle IoU=\frac{TP}{T+P-TP}\\
        \displaystyle \displaystyle nIoU=\frac{1}{N}\sum_i^N\frac{TP(i)}{T(i)+P(i)-TP(i)}\\
        \displaystyle \displaystyle  F_1=\frac{2TP}{2TP+FP+FN}
    \end{array}
\end{equation}
Where $N$ is the total number of samples, $T$ and $P$ denote the number of ground truth and predicted positive pixels, respectively. $TP$, $FP$, and $FN$ denote the number of true positive, false positive, and false negative pixels, respectively.
\subsection{Implementation Details}
We use SoftIoULoss as the loss function. AdamW is used as the optimizer and the poly learning rate decay strategy is used. And the deep supervision training strategy is used. We selected ACM \cite{dai2021asymmetric}, AGPCNet \cite{zhang2021agpcnet}, DNANet \cite{li2022dense}, RKFormer \cite{zhang2022rkformer}, ISNet \cite{zhang2022isnet}, IPI \cite{gao2013infrared}, RIPT \cite{dai2017reweighted}, and PSTNN \cite{zhang2019infrared} for comparison. Among them, IPI, RIPT, and PSTNN are traditional methods, and the rest are deep learning methods. We set the input dimension $C$ to 64 by default.

\begin{figure}[t]
    \centering
    \includegraphics[width=1\linewidth]{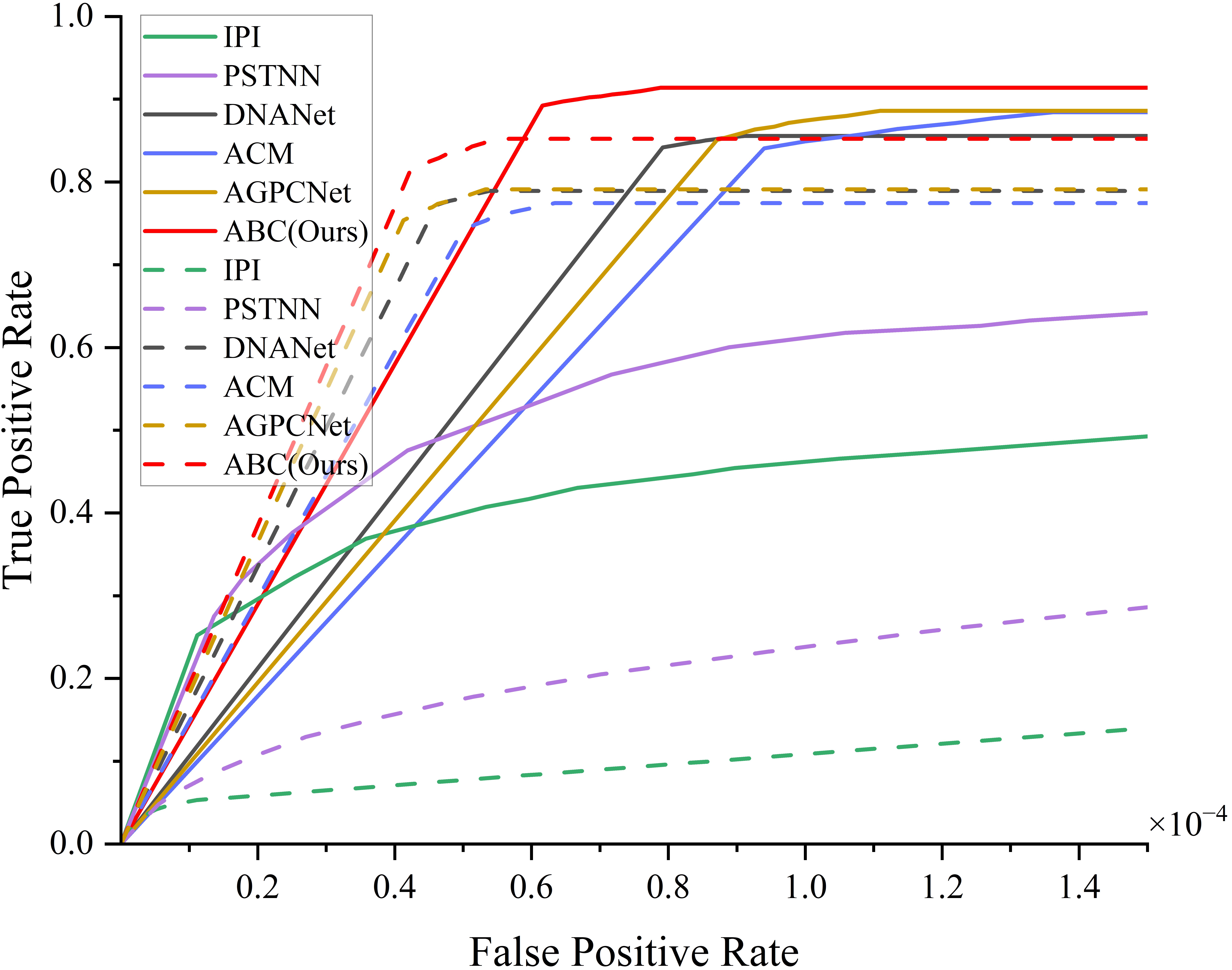}
    \caption{ROC curves of different methods on the NUAA dataset (solid line) and IRSTD1k dataset (dotted line).}
    \label{fig3}
\end{figure}

\subsection{Quantitative Results}
As shown in Table \ref{tab1}, on four datasets, our proposed ABC achieves the best performance in all metrics compared to other SOTA methods. In particular, on the IRSTD1k dataset, the $IoU$, $nIoU$, and $F_1$ of our method are 3.15\%, 1.28\%, and 2.09 higher than the second place, respectively. Where the traditional methods have lower metrics because most of the image backgrounds in these datasets are more complex, which leads to poor performance. Among the deep learning methods, the CNN based methods do not effectively handle noise and do not take into account the problem of deep target loss in the network, resulting in poor performance. Although RKFormer fuses transformer and CNN, it simply connects the two in parallel without deeper fusion, thus not achieving good performance. Our method deeply fuses CNN and transformer, allowing information interaction between local and global features, and the results show that it can effectively improve performance.

Fig. \ref{fig3} shows the ROC curves of different methods on the NUAA and IRSTD1k datasets. It can be seen that the performance of our method is significantly better than the other methods.

\begin{figure*}[ht]
    \centering
    \includegraphics[width=1\linewidth]{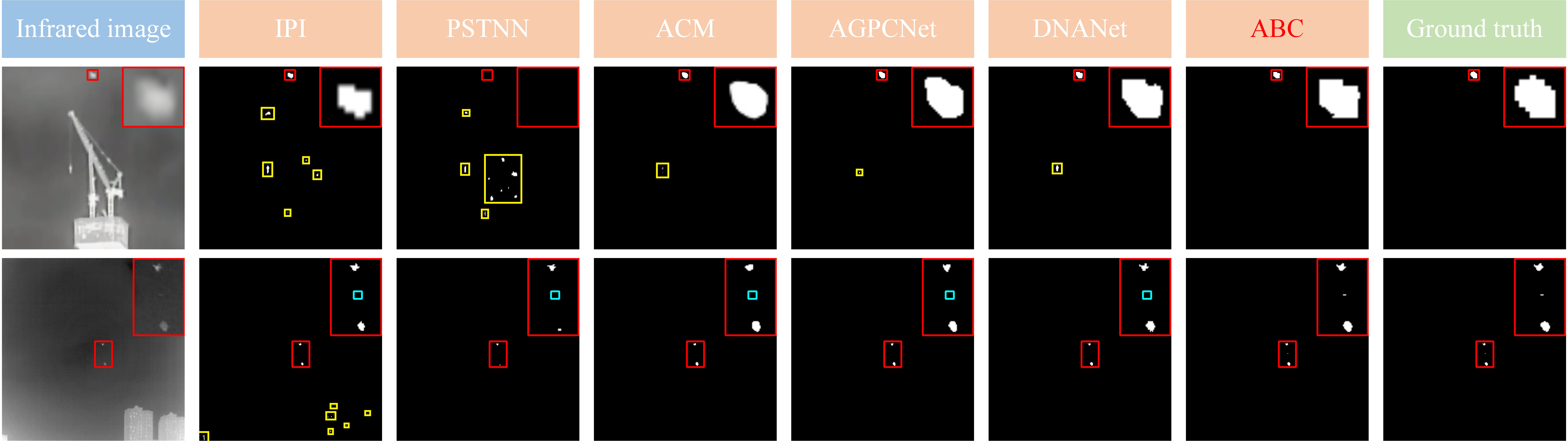}
    \caption{Partial image visualization results of different methods on NUAA and IRSTD1k datasets. The red box, the yellow box, and the cyan box represent the correct detection box, the false detection box, and the missed detection box, respectively.}
    \label{fig4}
\end{figure*}

\subsection{Visual Results}

Fig. \ref{fig4} shows the visualization results of some images from different methods on the NUAA and IRSTD1k datasets. It can be seen that the results of ABC are significantly better than the other methods. As shown in the first figure, other methods can easily produce false detections when IRST-like noise is present. But ABC benefits from the excellent noise suppression of the CLFT module and avoids false detection very well. Similar to the second figure, other methods are prone to miss detection when they encounter very small and faint target. Our method first goes through the CLFT module for a target feature enhancement to prevent target loss, then after refinement by the UCDC module, it can effectively detect very small and faint target and can make the whole segmentation result more accurate. More visualization results can be found in Sec. \ref{app_b} of the appendix.

\begin{table}[h]
\caption{The computational costs of different methods on NUAA dataset}
\tabcolsep=0.4cm
\renewcommand\arraystretch{1.5}
\begin{center}
\begin{tabular}{cccc}
\toprule[1pt]
Model  & FLOPs $\downarrow$ & FPS $\uparrow$  & IoU $\uparrow$ \\ \hline
ACM \cite{dai2021asymmetric}     & {\color[HTML]{FE0000} \textbf{1.14G}}   & {\color[HTML]{FE0000} \textbf{150}} & 72.88\%     \\
AGPCNet \cite{zhang2021agpcnet}  & 327.54G    & 17 & 77.13\%     \\
DNANet \cite{li2022dense} & 57.12G    & 49 & 74.91\%   \\ \hline
ABC-S(ours) & 21.26G    & 93  & 79.86\%   \\
ABC-B(ours)  & 83.75G    & 85 & 80.21\%   \\
ABC-L(ours)  & 332.55G   & 75 & {\color[HTML]{FE0000} \textbf{81.01\%}}    \\ \toprule[1pt]
\end{tabular}
\end{center}
\label{reflop}
\end{table}

\subsection{Performance Analysis}
The computational costs of different methods on NUAA dataset are shown in Table \ref{reflop}. The FPS was measured on a single RTX 3090. We categorized our model into small ($C=16$), base ($C=32$), and large ($C=64$) based on different input dimension $C$. It is evident that although ABC-L has a higher FLOPs, it also has the best performance and a higher FPS than AGPCNet \cite{zhang2021agpcnet} and DNANet \cite{li2022dense}. APC-S has excellent FLOPs and FPS, and its performance is also highly competitive.

\subsection{Ablation Study}
We conduct ablation studies on the NUAA dataset using the UNet model as a baseline. Related experiments on model design and hyperparameter settings can be found in Sec. \ref{app_c} and Sec. \ref{app_d} of the appendix. 

\textbf{Impact of CLFT Module:} As shown in Table \ref{tab2}, the second row is the result of removing the BAM from the CLFT module, it is equivalent to CLFT module retaining only the local feature extraction capability. The third row, on the contrary, is the result after retaining only the global feature extraction capability. Compared with the baseline, the performance is improved in general. However, none of them reach the performance of using the full CLFT module. The fourth row of results shows that using the full CLFT module has a large improvement in the performance of the model. It is shown that the CLFT module can effectively enhance the target features and suppress the noise.

\textbf{Impact of UCDC Module:} As shown in Table \ref{tab3}, when we use the UCDC module in UNet, the model performance is improved somewhat, indicating that the UCDC module can effectively handle the deep features of the network. The CLFT module is then added and the model performance is optimal, showing that the two can complement each other.

\begin{table}[t]
\caption{Ablation study of the CLFT module in $IoU(\%)$, $nIoU(\%)$, $F_1(10^{-2})$.}
\tabcolsep=0.30cm
\renewcommand\arraystretch{1.5}
\begin{center}
\begin{tabular}{cccc}
\toprule[1pt]
Method               & IoU $\uparrow$                                   & nIoU $\uparrow$                                  & $\rm F_1$ $\uparrow$                                    \\ \hline
UNet               &76.35                                       &77.93                                       & 86.59                                      \\
UNet+CLFT(only conv) &79.88                                       &78.38                                       &88.82                                       \\
UNet+CLFT(only BAM)   &77.97                                       & 77.31                                      & 87.62                                      \\
UNet+CLFT            & {\color[HTML]{FE0000} \textbf{80.17}} & {\color[HTML]{FE0000} \textbf{78.87}} & {\color[HTML]{FE0000} \textbf{89.00}} \\ \bottomrule[1pt]
\end{tabular}
\end{center}
\label{tab2}
\end{table}

\begin{table}[t]
\caption{Ablation study of the UCDC module in $IoU(\%)$, $nIoU(\%)$, $F_1(10^{-2})$.}
\tabcolsep=0.355cm
\renewcommand\arraystretch{1.5}
\begin{center}
    \begin{tabular}{cccc}
\toprule[1pt]
Method         & IoU $\uparrow$                                    & nIoU $\uparrow$                                   & $\rm F_1$ $\uparrow$                                     \\ \hline
UNet          & 76.35                                 & 77.93                                 & 86.59                                 \\
UNet+UCDC      & 79.75                                 & 78.05                                 & 88.73                                 \\
UNet+CLFT      & 80.17                                 & 78.87                                 & 89.00                                 \\
UNet+CLFT+UCDC & {\color[HTML]{FE0000} \textbf{81.01}} & {\color[HTML]{FE0000} \textbf{79.00}} & {\color[HTML]{FE0000} \textbf{89.51}} \\ \bottomrule[1pt]
\end{tabular}
\end{center}
\label{tab3}
\end{table}

\section{Conclusion}
In this paper, we propose ABC tailored towards the ISTD task, which addresses the challenges of feature loss and noise disturbance during transformations for small target. Concretely, we design two modules, CLFT and UCDC. The CLFT module employs a transformer architecture for local and global feature extraction, allowing the network to enhance the target features as well as suppress noise. The UCDC module serves to refine the output features of CLFT module further. Extensive experiments on public datasets validate our approach outperforms the SOTA methods.

\bibliographystyle{IEEEbib}
\bibliography{arXiv}

\appendix

\subsection{Experiment Details}
\label{app_a}
We trained four different models on four datasets. Due to differences in dataset distribution and resolution, we set different hyperparameters for each dataset. Due to space limitations, we did not provide detailed information in the main text. However, we created a configuration file for each dataset in our code, which readers can easily access and reproduce the experiments. Additional experimental details are provided in Table \ref{retab1}, in which the number of GPUs is 4, the scheduler uses the poly strategy, and none of the pretrained model is used.

\begin{table}[h]
\caption{Hyperparameter settings}
\tabcolsep=0.55cm
\renewcommand\arraystretch{1.5}
\begin{center}
\begin{tabular}{cccc}
\toprule[1pt]
Dataset  & Epochs & Lr     & Batch \\ \hline
NUAA \cite{dai2021asymmetric}     & 1500   & 0.0003 & 4     \\
IRSTD1k \cite{zhang2022isnet}  & 500    & 0.0001 & 4     \\
SIRSTAUG \cite{zhang2021agpcnet} & 500    & 0.0001 & 16    \\
NUDT \cite{li2022dense}     & 1500   & 0.0001 & 16    \\ \toprule[1pt]
\end{tabular}
\end{center}
\label{retab1}
\end{table}

\subsection{Visual Results}
\label{app_b}
\begin{figure*}[!t]
    \centering
    \includegraphics[width=1.0\linewidth]{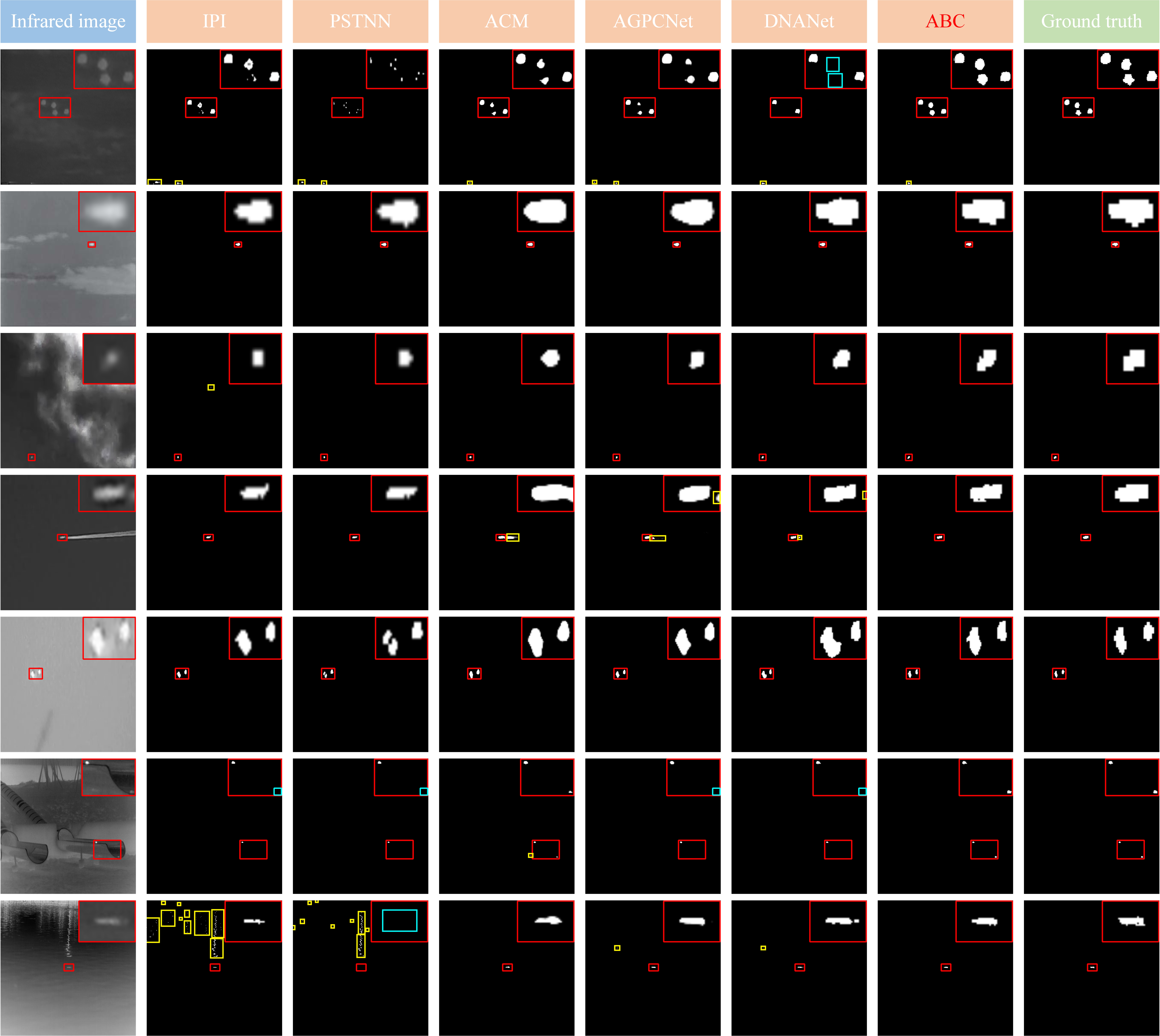}
    \caption{Partial image visualization results of different methods on NUAA and IRSTD1k datasets. The red box, the yellow box, and the cyan box represent the correct detection box, the false detection box, and the missed detection box, respectively.}
    \label{s_fig1}
\end{figure*}

Fig. \ref{s_fig1} shows the visualization results of different methods for some images on the NUAA \cite{dai2021asymmetric} and IRSTD1k \cite{zhang2022isnet} datasets. It can be seen that our method is still able to detect the target effectively even in the case of low contrast, complex background, and more noise interference. All other methods have different degrees of missed and false detection. Compared with our method, which not only has fewer missed and false detection but also can segment the targets better and the segmentation results are more fine compared with other methods.

\subsection{Model Design}
\label{app_c}
In this section, we investigate CLFT modules in the encoder part and UCDC modules in the decoder part. We have conducted experiments on the NUAA \cite{dai2021asymmetric} dataset using the standard ABC model as a baseline.

\textbf{Encoder:}
As shown in Table \ref{s_tab1}, we replaced the convolution module in the first layer of the encoder section with a CLFT module, which showed a significant degradation in performance. We believe that the infrared image has no clear semantic information and has a low signal-to-noise ratio. Since the CLFT module is dimensionally compressed when computing the attention matrix. If the complex original infrared image is directly used as the input, the CLFT module will be disturbed more when computing the attention matrix, and thus the wrong global information will be passed out. Therefore, we let the infrared image pass through a convolution module first, and the infrared image is coarsely extracted by the convolution module to filter out some background clutter so that the feature map input to the CLFT module does not have much interference. We find that the CLFT module has better results in processing the feature maps than the original infrared images directly.

\textbf{Decoder:}
As shown in Table \ref{s_tab2}, we replaced the UCDC module in the first layer of the decoder part with a convolution module, and its performance dropped slightly. We believe that when the feature map enters the decoder part through the transition layer, the resolution of the feature map is small. And after being processed by the UCDC module of the transition layer, the feature map is fine enough. The UCDC module has a larger receptive field than the convolution module, and it is more effective to process small-resolution feature maps than the convolution module. The two UCDC modules process feature maps at different scales, which can effectively filter out the noise interference around the target and make the target outline clearer.

\begin{table}[!t]
\caption{$IoU(\%)$, $nIoU(\%)$, $F_1(10^{-2})$ of whether use the ConvModule in the encoder.}
\tabcolsep=0.5cm
\renewcommand\arraystretch{1.5}
\begin{center}
    \begin{tabular}{cccc}
\toprule[1pt]
ConvModule         & IoU $\uparrow$                                    & nIoU $\uparrow$                                   & $\rm F_1$ $\uparrow$                                     \\ \hline
N           & 79.29                                 & 77.72                                 & 88.45                                 \\
Y & {\color[HTML]{FE0000} \textbf{81.01}} & {\color[HTML]{FE0000} \textbf{79.00}} & {\color[HTML]{FE0000} \textbf{89.51}} \\ \bottomrule[1pt]
\end{tabular}
\end{center}
\label{s_tab1}
\end{table}

\begin{table}[t]
\caption{$IoU(\%)$, $nIoU(\%)$, $F_1(10^{-2})$ of whether use the UCDC in the decoder.}
\tabcolsep=0.61cm
\renewcommand\arraystretch{1.5}
\begin{center}
    \begin{tabular}{cccc}
\toprule[1pt]
UCDC         & IoU $\uparrow$                                    & nIoU $\uparrow $                                  & $\rm F_1$ $\uparrow$                                     \\ \hline
N           & 79.82                                 & 77.19                                 & 88.78                                 \\
Y & {\color[HTML]{FE0000} \textbf{81.01}} & {\color[HTML]{FE0000} \textbf{79.00}} & {\color[HTML]{FE0000} \textbf{89.51}} \\ \bottomrule[1pt]
\end{tabular}
\end{center}
\label{s_tab2}
\end{table}

\subsection{Ablation Study}
\label{app_d}
In this section, we use the standard ABC model as a baseline and conduct ablation experiments on some hyperparameters in the model on the NUAA \cite{dai2021asymmetric} dataset.

\textbf{Impact of Input Dimension}
We study the impact of different input dimensions $C$ on the performance of the model, which is set to 64 by default in the paper. As shown in Table \ref{s_tab3}, when $C$ is set to 16, 32, and 64 respectively, the performance is also enhanced. But we found that when $C$ is set to 96, the performance drops slightly. We believe that since the infrared image has no clear semantic information, the inductive bias of the model will be affected when the input dimension is too large so that the model cannot effectively extract the target features.

\begin{table}[!t]
\caption{Ablation study of the input dimension in $IoU(\%)$, $nIoU(\%)$, $F_1(10^{-2})$.}
\tabcolsep=0.69cm
\renewcommand\arraystretch{1.5}
\begin{center}
    \begin{tabular}{cccc}
\toprule[1pt]
$C$         & IoU $\uparrow$                                    & nIoU $\uparrow$                                   & $\rm F_1$ $\uparrow$                                     \\ \hline
16           & 79.86                                 & 78.04                                 & 88.80                                 \\
32           & 80.21                                 & 78.15                                 & 89.02                                 \\
64 & {\color[HTML]{FE0000} \textbf{81.01}} & {\color[HTML]{FE0000} \textbf{79.00}} & {\color[HTML]{FE0000} \textbf{89.51}} \\ 
96           & 79.92                                 & 77.87                                 & 88.84                                 \\ \bottomrule[1pt]
\end{tabular}
\end{center}
\label{s_tab3}
\end{table}

\begin{table}[!t]
\caption{Ablation study of the dilated rate in $IoU(\%)$, $nIoU(\%)$, $F_1(10^{-2})$.}
\tabcolsep=0.5cm
\renewcommand\arraystretch{1.5}
\begin{center}
    \begin{tabular}{cccc}
\toprule[1pt]
Dilation Rate         & IoU $\uparrow$                                    & nIoU $\uparrow$                                   & $\rm F_1$ $\uparrow$                                     \\ \hline
1, 1, 1           & 80.12                                 & 77.55                                 & 88.96                                 \\
1, 2, 4           & 79.80                                 & 77.92                                 & 88.76                                 \\
2, 2, 2           & 80.91                                 & 78.40                                 & 89.45                                 \\
2, 4, 6           & 80.04                                 & 78.04                                 & 88.91                                 \\
2, 4, 2 & {\color[HTML]{FE0000} \textbf{81.01}} & {\color[HTML]{FE0000} \textbf{79.00}} & {\color[HTML]{FE0000} \textbf{89.51}} \\ \bottomrule[1pt]
\end{tabular}
\end{center}
\label{s_tab4}
\end{table}

\textbf{Impact of Dilation Rate:}
We study the effect of setting different dilation rates on the performance of the model in the three dilated convolutional layers in the CLFT module. As shown in Table \ref{s_tab4}, when we set the dilation rates of the three layers of dilated convolutional layers to the common 2, 4, and 6 respectively, the performance decreased slightly. The advantage of dilated convolution is that it has a larger receptive field, so when the dilation rate is set larger on conventional semantic segmentation tasks, it can bring more performance improvements. However, due to the small size of the infrared small target, using a larger dilation rate will introduce additional noise, resulting in blurred target features after feature fusion with the convolution branch. When the dilated rates are set too small, the receptive field will be relatively smaller, which will lead to the inability to effectively perceive the information around the target, resulting in performance degradation.  Therefore, setting the dilation rates to 2, 4, and 2 respectively can obtain long-distance information without introducing additional noise due to a large dilation rate.

\end{document}